\documentclass{article}

\PassOptionsToPackage{numbers}{natbib}
\usepackage[preprint]{neurips_2024}

\usepackage[table]{xcolor}
\usepackage{amsmath}
\usepackage{subcaption}
\usepackage{graphicx}
\usepackage{float}
\usepackage[utf8]{inputenc}
\usepackage[T1]{fontenc}
\usepackage{hyperref}
\usepackage{url}
\usepackage{booktabs}
\usepackage{amsfonts}
\usepackage{nicefrac}
\usepackage{microtype}
\usepackage{amssymb}
\usepackage{tikz}
\usetikzlibrary{arrows.meta}

\title{ICE-ID: A Novel Historical Census Dataset for Longitudinal Identity Resolution}

\author{%
   Gonçalo Hora de Carvalho\thanks{Corresponding author: \texttt{goncalo@iiim.is}} \\
  IIIM, Iceland \\
  \texttt{goncalo@iiim.is} \\
  \And
  Lazar S.~Popov \\
  IIIM, Iceland \\
  \And
  Sander Kaatee \\
  IIIM, Iceland \\
  \AND
  M\'ario S.~Correia \\
  IIIM, Iceland \\
  \AND
  Kristinn R.~Thórisson \\
  Full Research Professor, Department of Computer Science\\
  Reykjavik University\\
  \And
  Tangrui Li \\
  Temple University\\
  \And
  Pétur Húni Björnsson \\
  Department of Nordic Studies and Linguistics\\
  University of Copenhagen\\
  \And
  Eiríkur Smári Sigurðarson \\
  Center of Digital Humanities and Arts (CDHA)\\
  \And
  Jilles S.~Dibangoye \\
  Associate Professor, Machine Learning Group, Department of Artificial Intelligence\\
  Bernoulli Institute, University of Groningen\\
}

\begin{document}

\maketitle

\begin{abstract}
We introduce \textbf{ICE-ID}, a benchmark dataset comprising 984,028 records from 16 Icelandic census waves spanning 220 years (1703--1920), with 226,864 expert-curated person identifiers. ICE-ID combines hierarchical geography (farm$\to$parish$\to$district$\to$county), patronymic naming conventions, sparse kinship links (partner, father, mother), and multi-decadal temporal drift---challenges not captured by standard product-matching or citation datasets. This paper presents an artifact-backed analysis of temporal coverage, missingness, identifier ambiguity, candidate-generation efficiency, and cluster distributions, and situates ICE-ID against classical ER benchmarks (Abt--Buy, Amazon--Google, DBLP--ACM, DBLP--Scholar, Walmart--Amazon, iTunes--Amazon, Beer, Fodors--Zagats). We also define a deployment-faithful temporal OOD protocol and release the dataset, splits, regeneration scripts, analysis artifacts, and a dashboard for interactive exploration. Baseline model comparisons and end-to-end ER results are reported in the companion methods paper.
\end{abstract}


\section{Introduction}

Linking historical census records is fundamental to research on social mobility, demographic change, migration, and epidemiology, yet it remains arduous because names mutate, fields are missing, and administrative borders shift over time \cite{ruggles2018historical, bailey2020}. Most census-specific benchmarks oversimplify these challenges: they cover only short time ranges (often a single decade), omit kinship structure, and treat geography as flat text rather than a hierarchy \cite{papadakis2023}.

Carefully curated benchmarks can transform entire fields: \emph{ClimSim} unlocked hybrid physics--ML climate modelling \cite{ClimSim2024}; \emph{DecodingTrust} exposed safety gaps in frontier LLMs \cite{DecodingTrust2023}; the \emph{PRISM Alignment Dataset} broadened evaluation of alignment techniques across diverse regions \cite{PRISM2024}. Inspired by these successes, we release \textbf{ICE-ID}, the first large-scale open benchmark focused on \emph{long-term} person matching in a national population.

This paper focuses on the dataset itself: provenance, schema, statistical properties, and benchmark design for longitudinal identity resolution. Model evaluations and method comparisons are reported in the companion methods paper.

\textbf{Contributions.} We provide:
\begin{itemize}
    \item a longitudinal census dataset with stable identity labels, hierarchical geography, and optional kinship links;
    \item a temporal OOD evaluation protocol and task definitions suitable for both pairwise and clustering tracks;
    \item a set of dataset-centric diagnostics (temporal coverage, missingness, ambiguity, blocking efficiency, cluster-size CCDF) generated from published artifacts;
    \item a reproducible release with explicit provenance, licensing, a dashboard, and scripts that regenerate all tables and figures in this paper.
\end{itemize}


\section{Related Work}

\paragraph{Historical census linkage and record linkage datasets.}
Linking historical census records has long been central to research in social mobility, demography, migration, and epidemiology, yet it remains challenging due to name variation, missing or inconsistent attributes, and changes in administrative boundaries over time \cite{ruggles2018historical, Fellegi01121969, Newcombe1959AutomaticLO}. Prior work on census linkage has largely relied on proprietary or restricted-access data, often focusing on short temporal windows or single national censuses. While large-scale projects such as IPUMS provide extensive longitudinal census coverage, redistribution constraints and limited availability of downloadable record-level data restrict their use as open benchmarks \cite{sobek2011big, Sayers2015ProbabilisticRL}. Publicly available datasets that combine long temporal horizons with genealogical structure and explicit identity labels remain rare and expensive to produce.

\paragraph{Classical entity resolution benchmarks.}
Most widely used entity resolution (ER) benchmarks were developed for product matching or bibliographic deduplication tasks, including Abt-Buy, Amazon-Google, DBLP-ACM, and related datasets \cite{mudgal2018deep, Kpcke2010EvaluationOE}. These benchmarks are typically static snapshots, contain flat textual attributes, and lack temporal structure, demographic features, or relational context. As noted by Papadakis \emph{et al.} \cite{papadakis2023}, many such datasets are comparatively easy, with high performance achievable via simple similarity thresholds \cite{papadakis2023, Li2020DeepEM, Papadakis2019ASO}. As a result, they fail to capture the ambiguity, non-stationarity, and structural complexity characteristic of real-world population data.

\paragraph{Longitudinal and temporally structured benchmarks.}
Several recent benchmarks emphasize temporal or domain shift in tabular data, including TableShift \cite{gardner2023tableshift} and TabReD \cite{rubachev2024tabred}. These datasets demonstrate how distributional change can substantially affect model performance, but they do not include identity resolution tasks, genealogical relationships, or hierarchical geographic structure. Other longitudinal resources track identities over time in different domains---for example ORCID (researcher identities), parliamentary corpora such as ParlaMint, and correspondence corpora such as correspSearch---but they target different entity types and generally lack census-style demographic density, household structure, or kinship links \cite{orcid_about,orcid_public_data,erjavec2022parlamint,mueller2017correspsearch}.

\paragraph{Positioning of ICE-ID.}
ICE-ID complements existing work by providing an open benchmark for person-level identity resolution under long-term temporal drift. Unlike classical ER datasets, which are typically static and attribute-light, ICE-ID captures non-stationarity, name collisions, and structured geographic context in historical records. Unlike many longitudinal resources, which are restricted-access or centered on other entity types, ICE-ID is released with fixed temporal evaluation protocols and reproducible analysis artifacts. It therefore fills a gap between static ER benchmarks and large but non-redistributable census collections, enabling reproducible research on longitudinal identity resolution in realistic settings \cite{magellan}.


\section{Dataset}
\label{sec:dataset}

ICE-ID is a longitudinal census benchmark for person-level identity resolution under historical temporal drift. It contains \textbf{984,028 records} from \textbf{16 Icelandic census waves} spanning \textbf{1703--1920}, with \textbf{226,864} expert-curated person identifiers. Each row corresponds to one individual appearance in one census wave, allowing repeated observations across time.

The dataset captures several sources of real-world complexity that are largely absent from standard entity resolution benchmarks. Personal names follow Icelandic patronymic conventions, leading to systematic name collisions across the population. Demographic attributes such as birth year and marital status exhibit natural reporting noise and temporal inconsistency. Geographic information is encoded hierarchically (farm, parish, district, county) and evolves over time due to administrative boundary changes. Kinship links (partner, father, mother) are available but sparsely populated, reflecting historical census recording practices.

The release packages census appearances and expert-curated identity labels together with geographic tables encoding Iceland’s evolving administrative hierarchy. Ground-truth clusters are defined by the stable \texttt{person} identifier, enabling both pairwise linkage and multi-record clustering evaluation within and across census waves. This structure supports reproducible longitudinal ER experiments without relying on restricted-access census collections.

\subsection{Schema and Tables}

ICE-ID comprises four CSV files organized into two components:

\begin{enumerate}
    \item \textbf{Geographic tables} (\texttt{counties.csv}, \texttt{districts.csv}, \texttt{parishes.csv}): Encode Iceland's evolving territorial hierarchy. Each record carries a unique identifier, human-readable name, validity interval (``begins''/``ends''), and geographic centroids (\texttt{lat}, \texttt{lon}).
    
    \item \textbf{People table} (\texttt{people.csv}): 984,028 rows---one per individual appearance in each census wave (1703--1920). Each row captures:
    \begin{itemize}
        \item Name components: \texttt{nafn\_norm}, \texttt{first\_name}, \texttt{patronym}, \texttt{surname}
        \item Demographics: \texttt{birthyear}, \texttt{sex}, \texttt{status}, \texttt{marriagestatus}
        \item Cluster labels: \texttt{person} (expert-curated identity)
        \item Kinship links: \texttt{partner}, \texttt{father}, \texttt{mother} with provenance tags
        \item Geography: \texttt{farm}, \texttt{parish}, \texttt{district}, \texttt{county}
    \end{itemize}
\end{enumerate}


\section{Temporal Coverage and Label Density}
\label{sec:temporal_coverage}

\begin{figure}[ht]
  \centering
  \includegraphics[width=\linewidth]{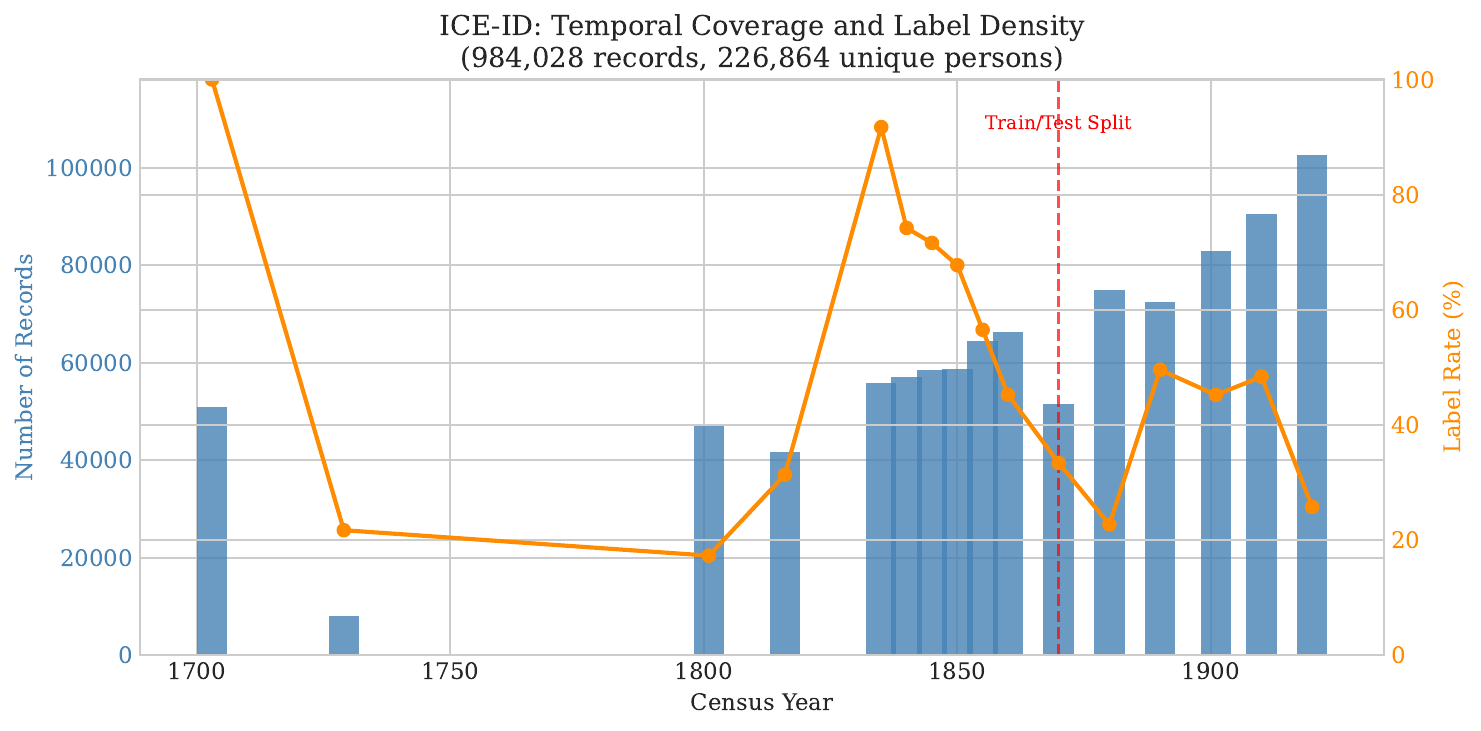}
  \caption{Temporal coverage and label density. ICE-ID spans 16 census waves; the 1703 census contains 50,959 records and the 1920 census contains 102,699. The average label rate (records with \texttt{person} assigned) is 50.17\%. Classic ER datasets are single-snapshot (``static'') and their label density is computed as the fraction of records appearing in at least one positive match pair.}
  \label{fig:temporal_coverage}
\end{figure}

Figure~\ref{fig:temporal_coverage} shows the distribution of records across census waves and the proportion with cluster labels.

 ICE-ID spans 16 census waves from 1703 to 1920. Record counts range from 8,072 (1729, partial) to 102,699 (1920). The total labeled population comprises 226,864 unique person identifiers, with 106,168 persons (46.8\%) appearing in multiple waves. The average label rate is 50.17\%.


\section{Missingness Over Time by Feature Family}
\label{sec:missingness}

\begin{figure}[ht]
  \centering
  \includegraphics[width=\linewidth]{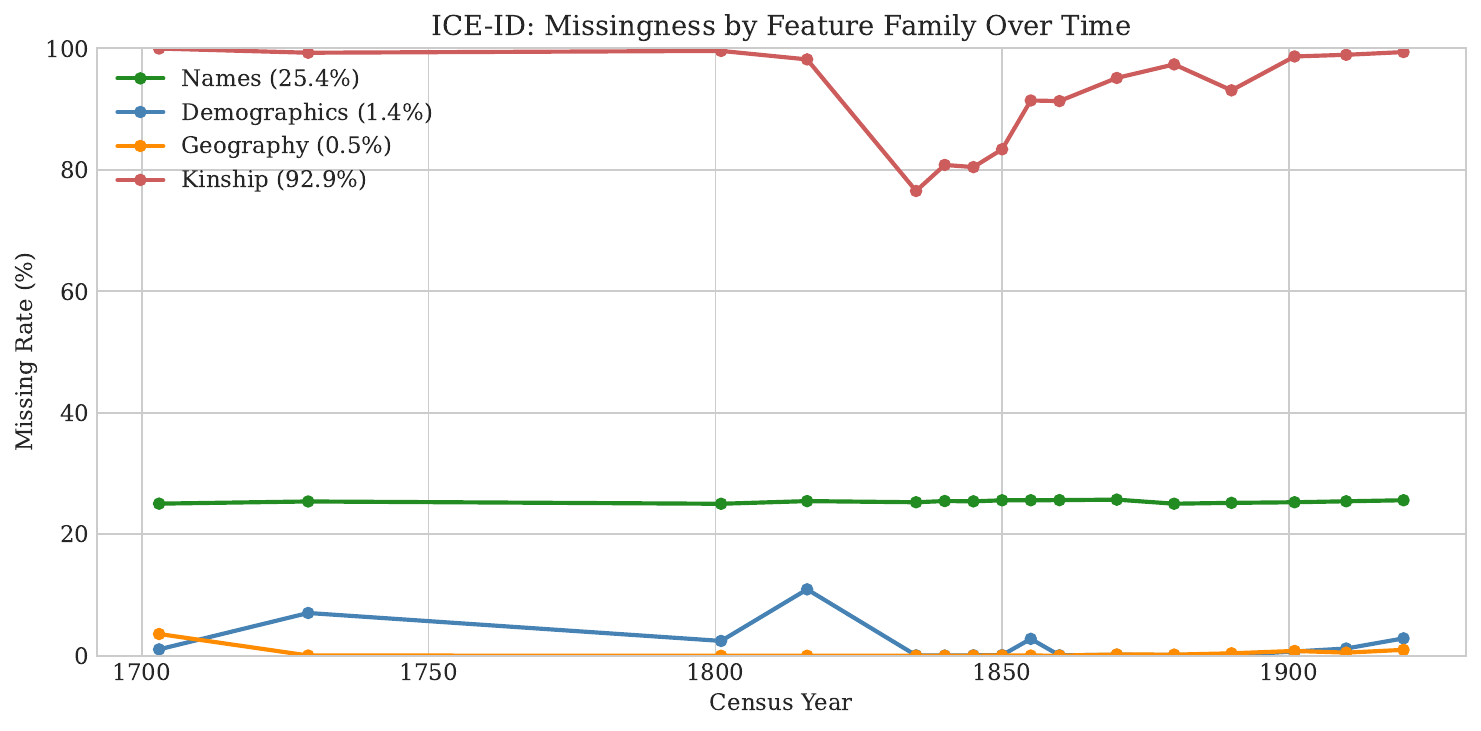}
  \caption{Missingness rates by feature family across census waves. Names are 25.39\% missing overall (dominated by surname), demographics are 1.38\% missing overall, geography is near-complete, and kinship links are 92.91\% missing overall.}
  \label{fig:missingness}
\end{figure}

Understanding missing data patterns is an important step prior to model development \cite{Naumann2014DataPR}. Figure~\ref{fig:missingness} shows missing-rate trajectories over time for four feature families.

Names are 25.39\% missing overall, driven primarily by surname. Demographics are 1.38\% missing overall (with most waves under 3\%). Kinship links are 92.91\% missing overall, explaining why methods that rely on household structure must be robust to sparse relational evidence.


\section{Entity Cluster Size Distribution}
\label{sec:cluster_sizes}

\begin{figure}[ht]
  \centering
  \includegraphics[width=0.8\linewidth]{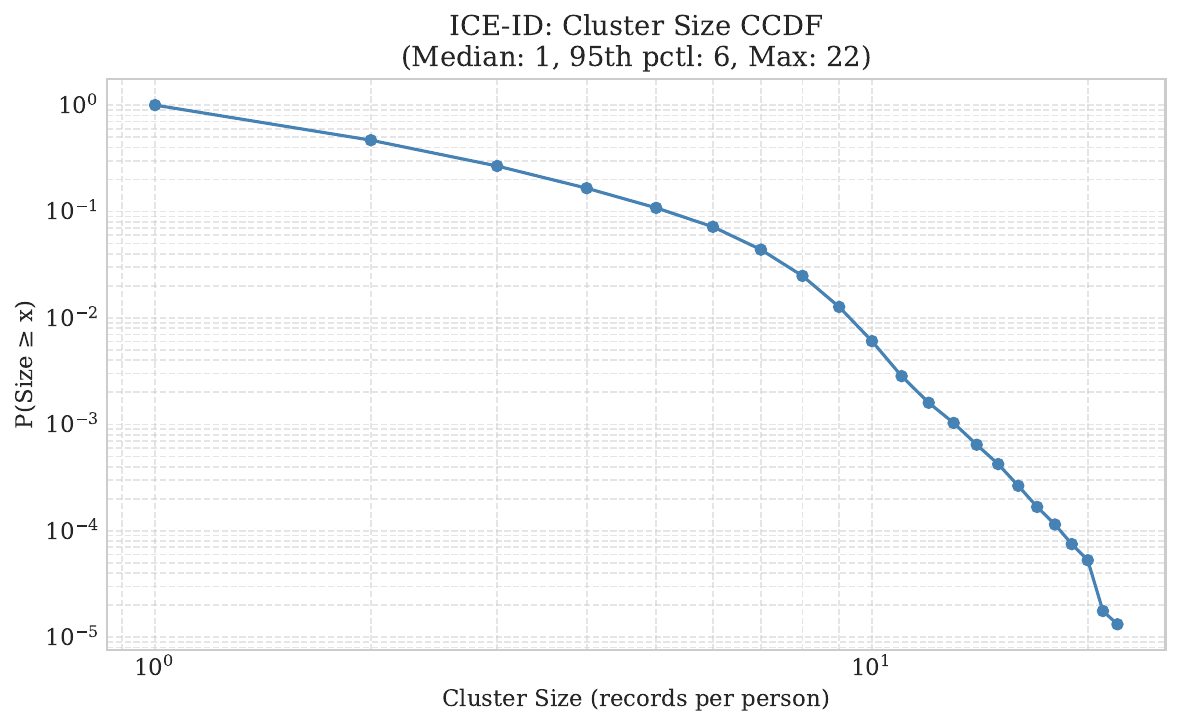}
  \caption{Cluster size CCDF (log--log). Median cluster size is 1; 95th percentile is 6; maximum cluster size is 22.}
  \label{fig:cluster_sizes}
\end{figure}

The cluster size distribution---how many census appearances per person---directly affects entity resolution difficulty. Figure~\ref{fig:cluster_sizes} shows the log--log CCDF\footnote{For cluster-size random variable $X \in \mathbb{N}$, the complementary cumulative distribution function is $F_c(k)=\Pr(X \ge k)=1-F(k-1)$, where $F$ is the CDF. Here, $F_c(k)$ is the proportion of person clusters with at least $k$ census appearances.}.

Exactly 120,696 persons (53.2\%) appear in only one census; 45,436 (20.0\%) appear in two. The maximum cluster size is 22 appearances. Median cluster size is 1; 95th percentile is 6. The long tail of large clusters (individuals appearing in 6+ censuses) represents high-value longitudinal subjects but also presents challenges for clustering algorithms that must maintain transitivity over many pairwise predictions.


\section{Identifier Ambiguity: Name Collision Analysis}
\label{sec:ambiguity}

\begin{figure}[ht]
  \centering
  \includegraphics[width=\linewidth]{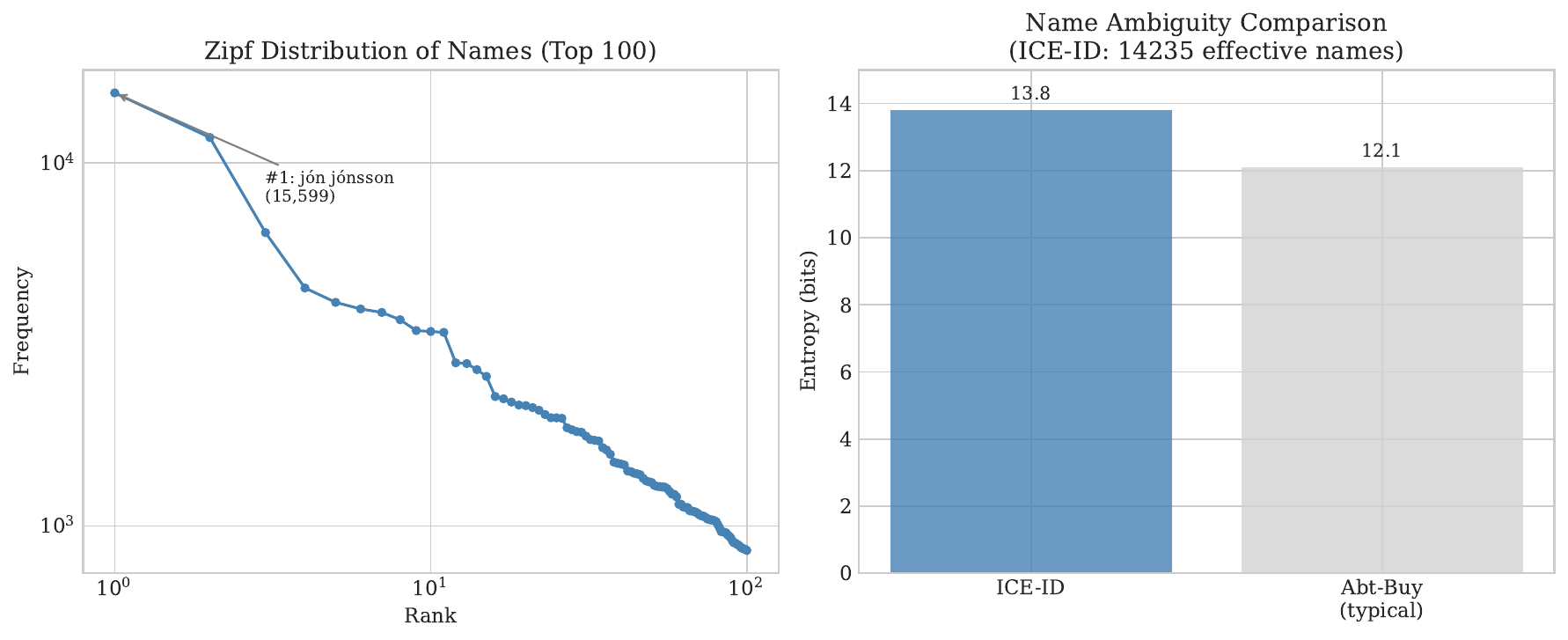}
  \caption{Name ambiguity analysis. (Left) Zipf plot of top 100 normalized names (\texttt{nafn\_norm}). (Right) Token entropy comparison between ICE-ID and representative classic ER datasets.}
  \label{fig:ambiguity}
\end{figure}

Patronymic naming conventions in Iceland create significant name collisions. Figure~\ref{fig:ambiguity} shows the Zipf distribution of normalized names and compares token entropy to classic ER datasets.

 The most common name (``Jón Jónsson'') appears 15,599 times. The top 10 names account for 6.2\% of all records. While the entropy (13.8 bits) is comparable to product datasets, the patronymic naming system creates systematic collisions: many individuals share identical names. This motivates the use of additional signals (birthyear, geography, kinship) for accurate disambiguation.


\section{Candidate Generation Efficiency}
\label{sec:blocking}

\begin{figure}[ht]
  \centering
  \includegraphics[width=0.8\linewidth]{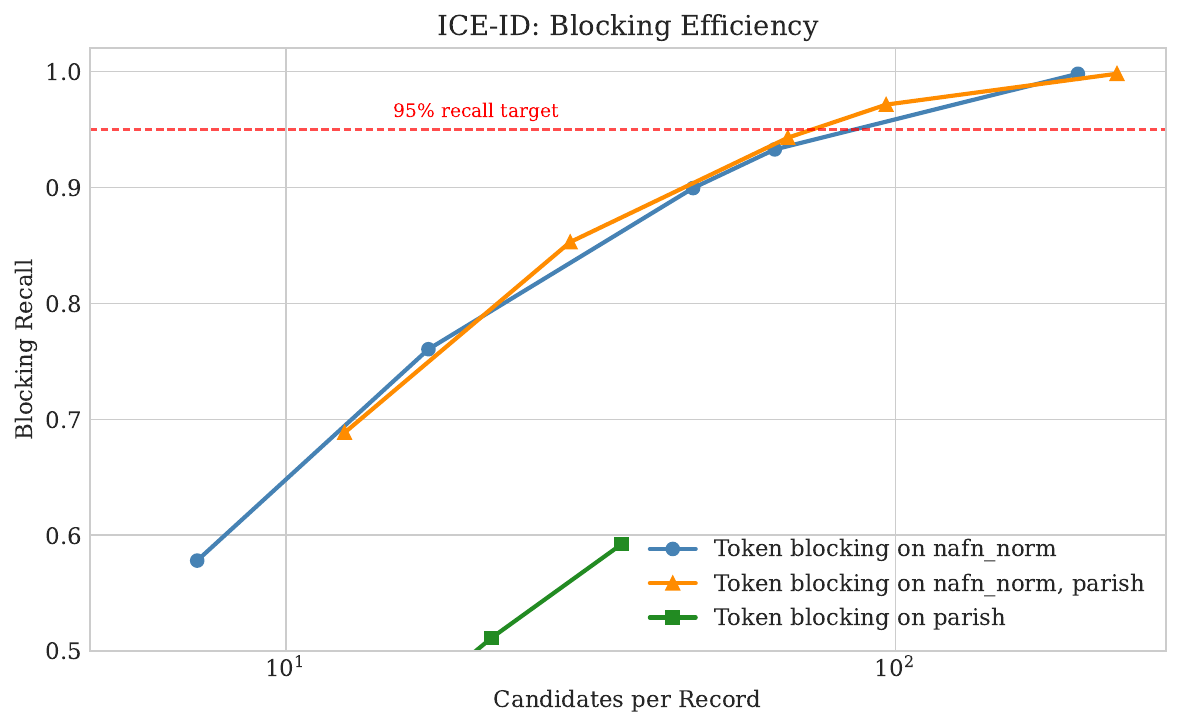}
  \caption{Blocking efficiency curves. Token blocking on \texttt{nafn\_norm} achieves 0.90 recall at 46.5 candidates/record; at 199 candidates/record it reaches 0.998 recall. Hybrid token blocking (name+parish) achieves 0.94 recall at 66.6 candidates/record and 0.97 recall at 96.5 candidates/record.}
  \label{fig:blocking}
\end{figure}

Scalable entity resolution requires effective blocking to reduce the $O(n^2)$ comparison space \cite{Christen2012ASO, Michelson2006LearningBS}. Figure~\ref{fig:blocking} shows blocking recall vs. candidate budget for different strategies.

 Pure geographic blocking (parish only) is insufficient due to population mobility and administrative boundary changes, achieving only 0.59 recall even with 35 candidates/record. Name-based blocking on \texttt{nafn\_norm} is more effective, reaching 0.90 recall at 46 candidates/record. The hybrid strategy (name + parish) provides the best recall/efficiency tradeoff, achieving 0.97 recall at 96.5 candidates/record; on this 7,466-record subset this corresponds to about 2.59\% of all possible unordered pairs.


\section{Missing Data Pattern Visualization}
\label{sec:missingno}

\begin{figure}[H]
  \centering
  \includegraphics[width=0.9\linewidth]{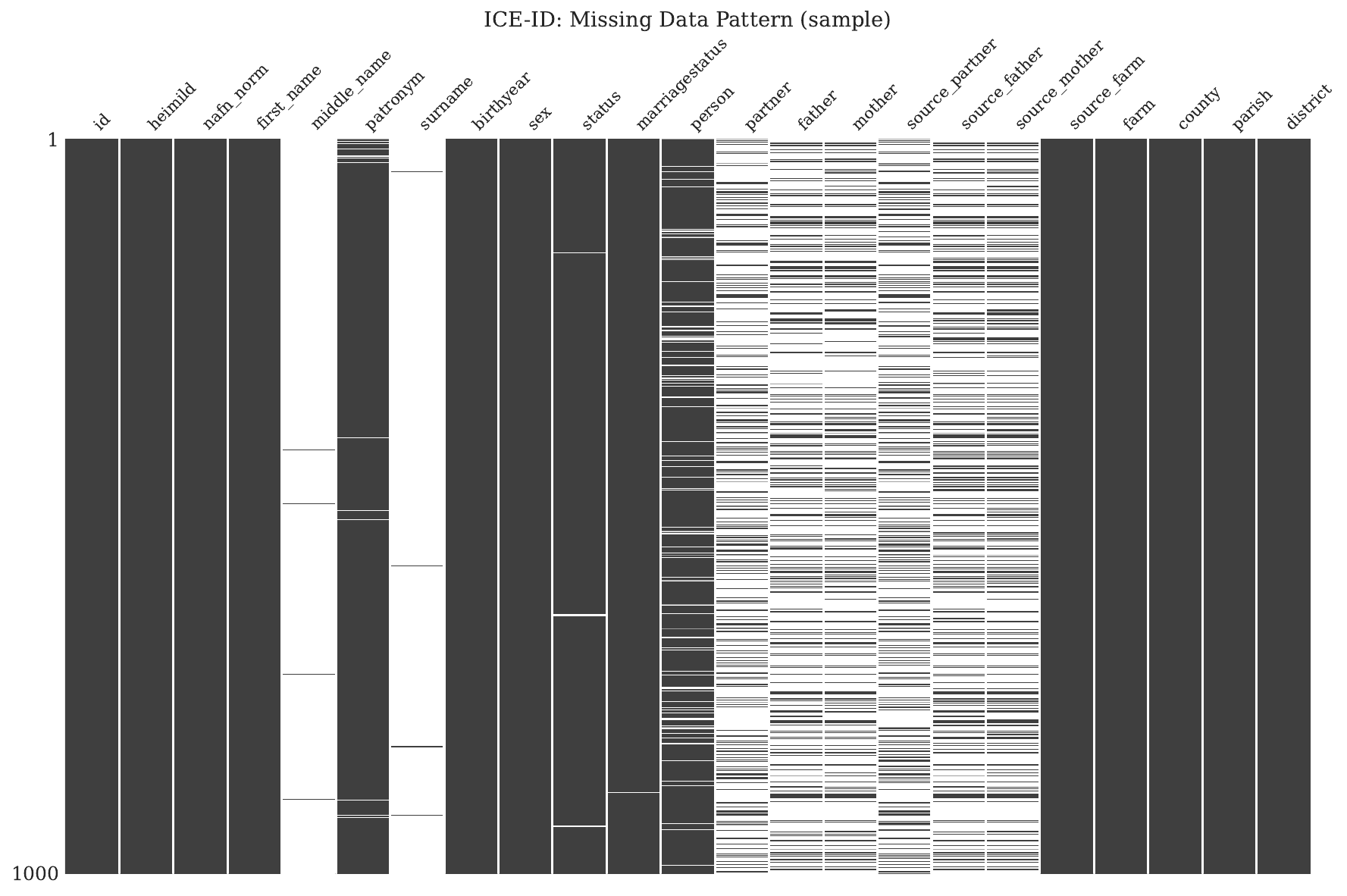}
  \caption{Missing data matrix (sample of 1,000 records). Geographic fields are near-complete; kinship links (partner, father, mother) are mostly missing; names show moderate missingness.}
  \label{fig:missingno}
\end{figure}

Figure~\ref{fig:missingno} provides a visual summary of missing data patterns in ICE-ID using a matrix plot. Each column represents a feature; white cells indicate missing values.

 The visualization confirms quantitative findings from Section~\ref{sec:missingness}: geographic hierarchy is reliably present, while kinship links are sparsely populated. It is simply the case that household relationships were inconsistently recorded, but mostly not at all.


\section{Comparison with Classical ER Benchmarks}
\label{sec:comparison}

Table~\ref{tab:dataset_synopsis} compares ICE-ID against standard entity resolution datasets.

\begin{table}[ht]
  \centering
  \caption{Dataset synopsis: ICE-ID vs. classical ER benchmarks.}
  \label{tab:dataset_synopsis}
  \scriptsize
  \begin{tabular}{lccccccc}
    \toprule
    \textbf{Dataset} & \textbf{Time Span} & \textbf{\#Waves} & \textbf{\#Records} & \textbf{\#Labeled IDs} & \textbf{\%Labeled} & \textbf{Geo} & \textbf{Kinship} \\
    \midrule
    ICE-ID & 1703--1920 & 16 & 984,028 & 226,864 & 50.2\% & Hierarchical & Yes \\
    Abt-Buy & N/A & 1 & 11,486 & 11,486 & 100\% & Flat & No \\
    Amazon-Google & N/A & 1 & 13,748 & 13,748 & 100\% & Flat & No \\
    DBLP-ACM & N/A & 1 & 14,834 & 14,834 & 100\% & Flat & No \\
    DBLP-Scholar & N/A & 1 & 34,446 & 34,446 & 100\% & Flat & No \\
    Walmart-Amazon & N/A & 1 & 12,288 & 12,288 & 100\% & Flat & No \\
    iTunes-Amazon & N/A & 1 & 642 & 642 & 100\% & Flat & No \\
    Beer & N/A & 1 & 536 & 536 & 100\% & Flat & No \\
    Fodors-Zagats & N/A & 1 & 1,134 & 1,134 & 100\% & Flat & No \\
    \bottomrule
  \end{tabular}
\end{table}

Table~\ref{tab:schema_matrix} provides a schema comparability matrix showing feature availability across datasets.

\begin{table}[ht]
  \centering
  \caption{Schema comparability matrix. \checkmark = present, $\sim$ = partial, --- = absent.}
  \label{tab:schema_matrix}
  \scriptsize
  \begin{tabular}{lccccccccc}
    \toprule
    \textbf{Feature Family} & \textbf{ICE-ID} & \textbf{Abt-Buy} & \textbf{Amz-Ggl} & \textbf{DBLP-ACM} & \textbf{Wlm-Amz} & \textbf{iTun-Amz} & \textbf{Beer} & \textbf{Fod-Zag} \\
    \midrule
    Name / Title & \checkmark & \checkmark & \checkmark & \checkmark & \checkmark & \checkmark & \checkmark & \checkmark \\
    Age / Birthyear & \checkmark & --- & --- & --- & --- & --- & --- & --- \\
    Sex / Gender & \checkmark & --- & --- & --- & --- & --- & --- & --- \\
    Household / Family & \checkmark & --- & --- & --- & --- & --- & --- & --- \\
    Parent links & \checkmark & --- & --- & --- & --- & --- & --- & --- \\
    Spouse / Partner & \checkmark & --- & --- & --- & --- & --- & --- & --- \\
    Address / Geo & \checkmark (4-level) & --- & --- & --- & --- & --- & --- & \checkmark \\
    Temporal field & \checkmark & --- & --- & $\sim$ (year) & --- & $\sim$ (released) & --- & --- \\
    Free-text notes & $\sim$ & \checkmark & \checkmark & --- & \checkmark & --- & --- & --- \\
    \bottomrule
  \end{tabular}
\end{table}

 ICE-ID is the only dataset combining temporal coverage, hierarchical geography, and kinship signals. Classical ER benchmarks lack demographic attributes entirely and provide only flat text fields for matching. This makes ICE-ID uniquely suited for evaluating methods that leverage structured signals.


\section{Comparison with Longitudinal Datasets}
\label{sec:longitudinal}

While classical ER benchmarks are static snapshots, several other datasets capture identity over time. Table~\ref{tab:longitudinal} compares ICE-ID to longitudinal identity datasets.

\begin{table}[ht]
  \centering
  \caption{Longitudinal dataset comparison. ``File'' = downloadable data; ``Doc'' = metadata only.}
  \label{tab:longitudinal}
  \scriptsize
  \begin{tabular}{lcccccc}
    \toprule
    \textbf{Dataset} & \textbf{Time Span} & \textbf{Entity Type} & \textbf{$\sim$Entities} & \textbf{Temporal Signal} & \textbf{Data} & \textbf{Access} \\
    \midrule
    ICE-ID & 1703--1920 & Person & 227K & Census year & File & Open \\
    IPUMS LRS & 1850--1940 & Person & 50M & Census year & Doc & Account \\
    IPUMS MLP & 1870--2020 & Person & 100M & Census+survey & Doc & Account \\
    IPUMS NAPP & 1801--1910 & Person & 100M & Census year & Doc & Account \\
    ORCID & 2012--present & Researcher & 18M & Last modified & File (sample) & Open \\
    SemParl & 1907--2021 & Parliamentarian & 7K & Speech date & File & Open \\
    CKCC & 1600--1800 & Correspondent & 5K & Letter date & File & Open \\
    correspSearch & 1500--2000 & Correspondent & 130K letters & Letter date & File & Open \\
    Synthea & 1950--2020 & Patient & 1K (sample) & Encounter date & File & Open \\
    FEBRL & N/A & Patient & 5--10K & None (static) & File & Open \\
    \bottomrule
  \end{tabular}
\end{table}

 IPUMS datasets provide the largest historical census coverage but require account access and do not function as fully redistributable ER benchmarks. Open alternatives (ORCID, parliamentary corpora, correspondence KGs) cover different entity types and temporal signals. ICE-ID combines a multi-century census setting with hierarchical geography and kinship fields in an openly downloadable benchmark release.

\section{Benchmark Tasks and Protocols}
\label{sec:tasks}

ICE-ID supports evaluation of identity resolution methods under realistic temporal distribution shift. We define two canonical tasks and a strictly temporal evaluation protocol intended to reflect deployment settings where early census waves are available for training while later waves are held out for testing.

\subsection{Tasks}
We define two benchmark tasks:
\begin{enumerate}
    \item \textbf{Within-wave (intra-census) linkage:} determine whether two records from the same census wave refer to the same individual.
    \item \textbf{Across-wave (cross-census) linkage:} determine whether two records from different census waves refer to the same individual, despite multi-decadal drift in name spelling, reported birthyear, residence, and administrative boundaries.
\end{enumerate}
Unless otherwise specified, cross-census evaluation is performed on records drawn from adjacent census waves within a split; the protocol can also be applied to non-adjacent waves to probe longer-range temporal drift.

\subsection{Temporal splits}
We propose the following temporal split:
\begin{itemize}
    \item \textbf{Train:} pre-1870 (560,334 records; 153,311 unique persons)
    \item \textbf{Validation:} 1871--1890 (147,450 records; 44,645 unique persons)
    \item \textbf{Test:} 1891--1920 (276,244 records; 62,109 unique persons)
\end{itemize}
This split induces a temporal out-of-distribution (OOD) setting in which the model must generalize from earlier censuses to later ones. A small cross-split overlap exists: 6,136 persons (2.7\% of unique persons) appear in both train and test splits, enabling direct evaluation of temporal generalization on individuals observed across long time spans.

\subsection{Ground truth and pair construction}
Ground-truth identities are defined by the expert-curated cluster label \texttt{person}. A \textbf{positive pair} is any pair of records within the evaluation split that share the same \texttt{person} identifier. \textbf{Negative pairs} are sampled from records with different \texttt{person} identifiers. To limit trivial negatives and reflect realistic candidate sets, negatives are sampled \emph{within the same blocking partition} (e.g., as produced by a candidate-generation strategy), using a default ratio of two negatives per positive \cite{payasyougo}.

\subsection{Evaluation modes and metrics}
ICE-ID supports evaluation in both \textbf{pairwise} and \textbf{clustering} modes.
\begin{itemize}
    \item \textbf{Pairwise evaluation} scores predicted match decisions on labeled pairs using standard metrics such as precision, recall, and F$_1$.
    \item \textbf{Clustering evaluation} assesses the quality of predicted identity clusters against ground-truth \texttt{person} clusters using clustering metrics such as Adjusted Rand Index (ARI).
\end{itemize}
For clustering-based systems, outputs should respect \textbf{transitivity}: if record $A$ matches $B$ and $B$ matches $C$, then $A$ and $C$ should be placed in the same predicted entity cluster. Reporting both pairwise and clustering metrics is recommended, as high pairwise accuracy does not necessarily imply globally coherent clusters under transitivity constraints.


\subsection{Protocol Summary}
\label{sec:protocols_splits}

Table~\ref{tab:protocols} summarizes the canonical evaluation protocol for ICE-ID.

\begin{table}[ht]
  \centering
  \caption{Evaluation protocols and temporal splits.}
  \label{tab:protocols}
  \small
  \begin{tabular}{lp{10cm}}
    \toprule
    \textbf{Component} & \textbf{Specification} \\
    \midrule
    \textbf{Temporal splits} & Train: up to 1870 (560,334 records, 153,311 unique persons); Val: 1871--1890 (147,450 records, 44,645 persons); Test: 1891--1920 (276,244 records, 62,109 persons) \\
    \textbf{Positive pairs} & All pairs of records sharing the same \texttt{person} ID within the evaluation split \\
    \textbf{Negative sampling} & 2 negatives per positive, sampled from same blocking partition \\
    \textbf{Transitivity} & Ground-truth clusters defined by \texttt{person} field; methods should enforce transitivity in clustering outputs \\
    \textbf{Evaluation modes} & Within-wave (same census) and cross-wave (adjacent censuses) \\
    \bottomrule
  \end{tabular}
\end{table}



\section{Dataset Card}
\label{sec:dataset_card}

Following the Datasheets for Datasets framework \cite{gebru2021datasheets, Bender2018DataSF, Mitchell2018ModelCF}:

\subsection{Motivation and Intended Use}
\textbf{Purpose:} Provide a realistic, large-scale benchmark for longitudinal identity resolution.

\textbf{Intended Uses:} (1) Entity resolution research; (2) Temporal distribution shift studies; (3) Genealogical/historical research.

\textbf{Out-of-Scope Uses:} (1) Re-identification of living individuals (data ends 1920); (2) Commercial genealogy without attribution.

\subsection{Composition}
\textbf{Records:} 984,028 individual census appearances.
\textbf{Labeled Entities:} 226,864 unique person clusters (106,168 with multiple records).
\textbf{Fields:} 23 columns including names, demographics, kinship, geography.

\subsection{Collection and Labeling}
\textbf{Collection:} Digitized from Icelandic census manuscripts by National Archives.
\textbf{Labels:} Expert-curated using genealogical records and parish registers.

\subsection{Maintenance}
\textbf{License:} CC-BY-4.0. \\
\textbf{Dataset Access:} \url{https://huggingface.co/datasets/goldpotatoes/ice-id} \\
\textbf{Code and Artifacts:} \url{https://github.com/IIIM-IS/ICE-ID-2.0}

\subsection{Interactive Exploration and Model Tooling}
\label{sec:dashboard}

To facilitate engagement with the ICE-ID ecosystem, we provide an interactive Streamlit-based dashboard for end-to-end entity resolution and data exploration (see Figure~\ref{fig:dashboard_preview}). The application is organized into four primary workspaces:

\begin{figure}[H]
     \centering
     \begin{subfigure}[t]{0.46\textwidth}
         \centering
         \includegraphics[width=\linewidth]{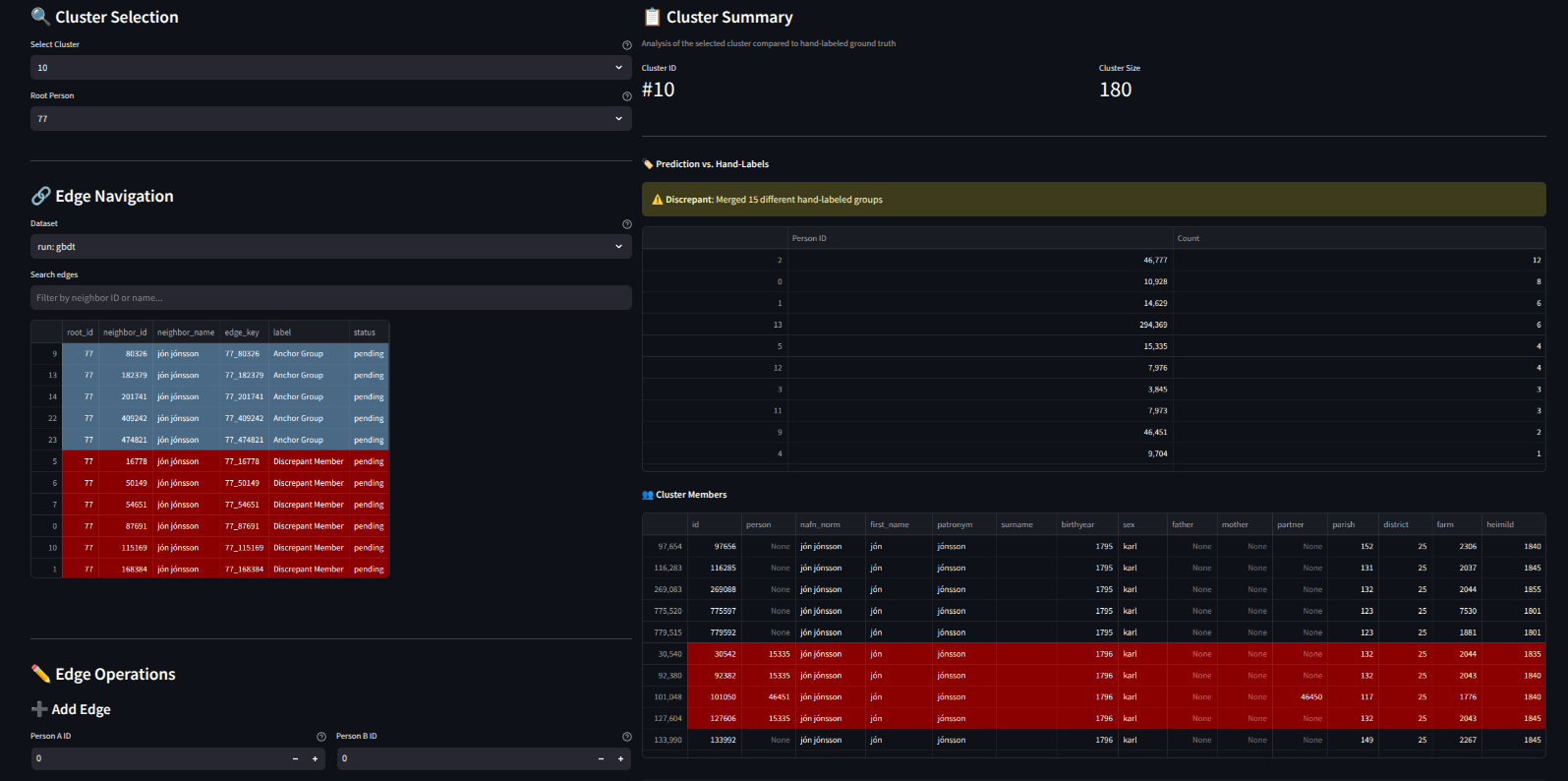}
         \caption{Cluster browsing interface showing edge navigation, prediction-vs-ground-truth comparison, and colour-coded member details for a selected cluster.}
         \label{fig:dash_tables}
     \end{subfigure}
     \hfill
     \begin{subfigure}[t]{0.51\textwidth}
         \centering
         \includegraphics[width=\linewidth]{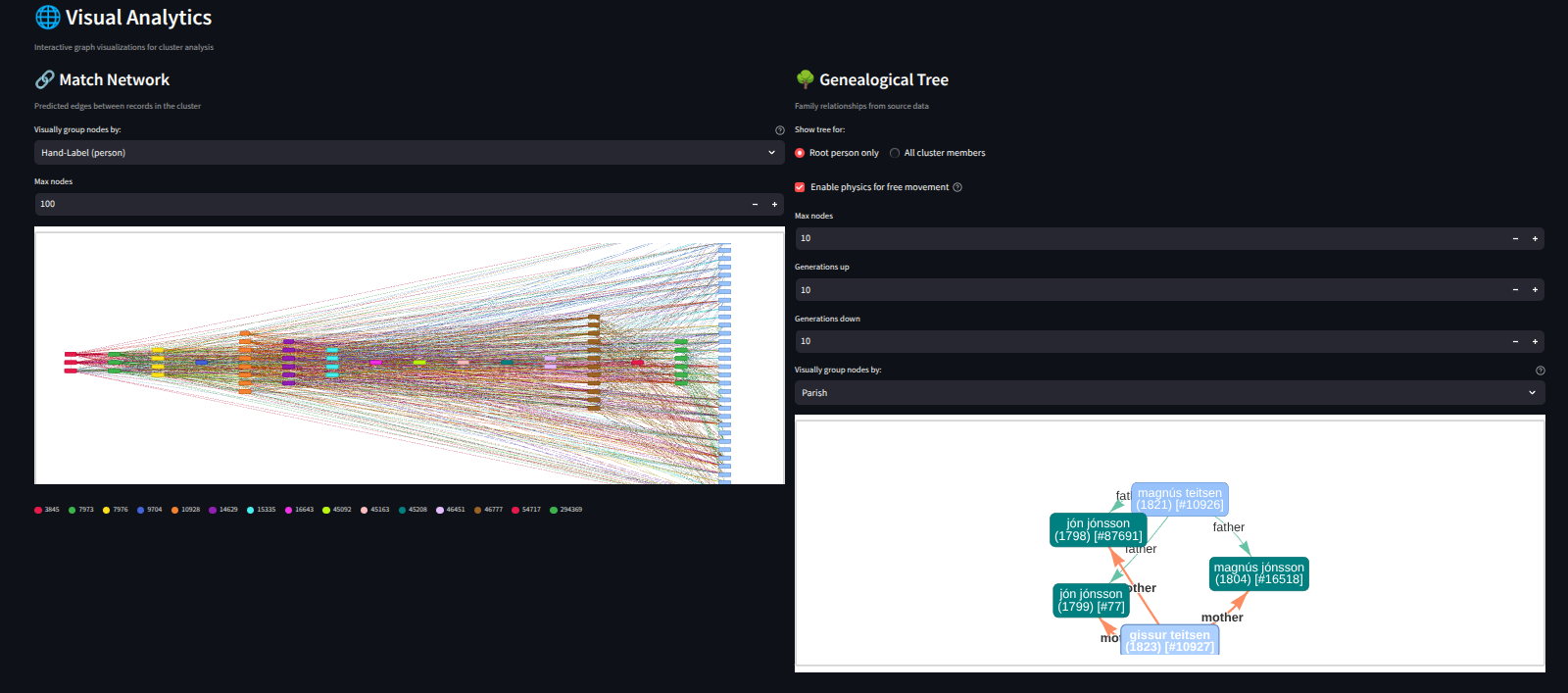}
         \caption{Interactive match network of predicted edges (left) and genealogical tree of family relationships (right) for visual validation of linked records.}
         \label{fig:dash_networks}
     \end{subfigure}
     \caption{Dashboard - two example views.}
     \label{fig:dashboard_preview}
\end{figure}

\begin{itemize}
    \item \textbf{Training:} Supports the configuration and execution of the ICE-ID pipeline alongside 12 external baselines (e.g., \textit{Ditto}, \textit{XGBoost}, \textit{Splink}) with real-time log streaming and hyperparameter management.
    \item \textbf{Evaluation:} Computes pairwise metrics with Wilson confidence intervals and cluster-level statistics (weighted purity, singleton rates), generating automated interpretive commentary for results.
    \item \textbf{Inspect \& Edit:} A debugging interface featuring \texttt{pyvis}-based match networks and genealogical trees. It allows for live edge manipulation with automatic Union-Find cluster recomputation and a feedback loop for model finetuning via JSON-based label exports.
    \item \textbf{Data Editor:} Provides tools for reshaping the underlying CSVs, visualizing missing-data patterns (matrix/correlation plots), and managing persistent model configuration profiles.
\end{itemize}


\section{Limitations and Ethical Considerations}
\label{sec:limitations}

\textbf{Limitations:}
\begin{itemize}
    \item \textbf{Snapshot waves only:} Continuous life-course trajectories are inferred, not observed.
    \item \textbf{Kinship sparsity:} Only 7\% of records have partner/parent links.
    \item \textbf{Label noise:} Unknown amount of transcription error rates.
    \item \textbf{Coverage gaps:} 1729 and 1870 censuses are partial.
\end{itemize}

\textbf{Ethical considerations:} Two facts minimize ethical issues: firstly, the family-link records are sparsely populated such that many persons lack explicit relational connections to other persons, for example, parental relationship or marital counterparts. Secondly, census data ends in 1920, so no direct link to living individuals exists. The dataset enables historical/demographic research while minimizing re-identification risks.


\section{Conclusion}

ICE-ID provides a unique resource for entity resolution research: a century-spanning, genealogically enriched dataset with hierarchical geography and temporal drift. Its structure exposes challenges (name ambiguity, missingness, cluster size heterogeneity) absent from classical benchmarks. By releasing data, protocols, and analysis artifacts, we enable reproducible research on longitudinal identity resolution. For model evaluations and method comparisons, see our companion paper.

\section*{Acknowledgements}

We acknowledge the \textbf{Center of Digital Humanities and Arts (CDHA), Iceland}, for enabling access to and supporting the use of the historical census and registry data underlying ICE-ID. We thank the technical and administrative staff involved in the digitization, preservation, and curation of these records, whose work made this dataset available for research. We also acknowledge institutional support from the participating universities and research centers that facilitated data processing, analysis, and the open release of this benchmark.


\bibliographystyle{plainnat}
\bibliography{main}


\appendix

\section{Dataset Access}
Full dataset available at \url{https://huggingface.co/datasets/goldpotatoes/ice-id}.

\section{Supplementary Statistics}
Detailed per-census statistics and additional analysis artifacts are provided in the repository.

\section{Artifact Provenance Graph}

\begin{figure}[ht]
  \centering
  \resizebox{\linewidth}{!}{%
  \begin{tikzpicture}[
    node distance=10mm and 15mm,
    >=Latex,
    box/.style={draw, rounded corners, align=left, font=\small, inner sep=4pt, text width=5.5cm},
    title/.style={font=\bfseries\small}
  ]
    \node[title] (c1) at (0, 3.5) {Data Sources};
    \node[title] (c2) at (6.5, 3.5) {Generation Script};
    \node[title] (c3) at (15.0, 3.5) {Artifacts};
    \node[title] (c4) at (23.0, 3.5) {Paper Outputs};

    \node[box, text width=5cm] (d1) at (0, 2.0) {\texttt{raw\_data/people.csv}};
    \node[box, text width=5cm] (d2) at (0, 0.4) {\texttt{bench/deepmatcher\_data/*}};
    \node[box, text width=5cm] (d3) at (0, -1.2) {Curated longitudinal reference values};

    \node[box, text width=6cm] (s1) at (6.5, 0.4) {\texttt{bench/scripts/}\\ \texttt{generate\_paper\_artifacts.py}};

    \node[box, text width=7.5cm] (a1) at (15.0, 2.0) {\texttt{bench/paper\_artifacts/plot\_data/}\\ \texttt{fig1\_temporal\_coverage.json}, \texttt{fig2\_missingness.json}, \texttt{fig3\_cluster\_sizes.json}, \texttt{fig4\_ambiguity.json}, \texttt{fig5\_blocking.json}};
    \node[box, text width=7.5cm] (a2) at (15.0, -0.4) {\texttt{bench/paper\_artifacts/table\_data/}\\ \texttt{table1\_dataset\_synopsis.csv}, \texttt{table2\_schema\_matrix.csv}, \texttt{table3\_protocols\_splits.csv}, \texttt{table\_longitudinal\_comparison.csv}};
    \node[box, text width=7.5cm] (a3) at (15.0, -2.4) {\texttt{papers/figures/fig1--fig5.pdf}\\ \texttt{papers/figures/missingno\_iceid.pdf}};

    \node[box, text width=6cm] (p1) at (23.0, 1.2) {Figures~\ref{fig:temporal_coverage}--\ref{fig:blocking}, Figure~\ref{fig:missingno}};
    \node[box, text width=6cm] (p2) at (23.0, -1.4) {Tables~\ref{tab:dataset_synopsis}, \ref{tab:schema_matrix}, \ref{tab:longitudinal}, \ref{tab:protocols}};

    \draw[->] (d1) -- (s1.north west);
    \draw[->] (d2) -- (s1.west);
    \draw[->] (d3) -- (s1.south west);
    \draw[->] (s1.north east) -- (a1.west);
    \draw[->] (s1.east) -- (a2.west);
    \draw[->] (s1.south east) -- (a3.west);
    \draw[->] (a1.east) -- (p1.west);
    \draw[->] (a3.east) -- (p1.south west);
    \draw[->] (a2.east) -- (p2.west);
  \end{tikzpicture}%
  }
  \caption{Data-paper provenance graph mapping source data, generation script, saved artifacts, and paper figures/tables.}
  \label{fig:data_provenance_graph}
\end{figure}

\end{document}